\documentclass[conference,a4paper]{APSIPA2026}
\usepackage{amsmath}
\usepackage{graphicx}
\usepackage{multirow}
\usepackage{threeparttable}
\usepackage{kotex}
\usepackage{stfloats}
\usepackage{makecell}
\usepackage[backend=biber,style=ieee,]{biblatex}
\usepackage{twemojis}
\usepackage{booktabs}
\DeclareRobustCommand{\cofirstmark}{\raisebox{0.7ex}{\twemoji[height=1.1ex]{taco}}}

\usepackage{geometry}
\usepackage{fancyhdr}

\fancypagestyle{firststyle}{
  \fancyhf{}
  \fancyhead[C]{2026 Asia Pacific Signal and Information Processing Association Annual Summit and Conference (APSIPA ASC)}
}

\begin{document}

\title{Analyzing Error Propagation in\\Korean Spoken QA with ASR--LLM Cascades}
\author{
\authorblockN{
Donghyuk Jung\authorrefmark{1}\cofirstmark and Youngwon Choi\authorrefmark{2}\cofirstmark }

\authorblockA{
\authorrefmark{1}
Korea Culture Technology Institute, Republic of Korea \\
E-mail: dhjung081121@gm.gist.ac.kr
}

\authorblockA{
\authorrefmark{2}
Maum AI Inc., Republic of Korea \\
E-mail: youngwonchoi@maum.ai
}

}

\maketitle
\begingroup
\renewcommand{\thefootnote}{}
\footnotetext{\cofirstmark{} These authors contributed equally to this work. This work was conducted as a collaborative research activity within De Samo, an independent research~group.
}
\endgroup
\thispagestyle{firststyle}
\pagestyle{empty}

\begin{abstract}
We analyze how automatic speech recognition (ASR) errors propagate through ASR--LLM cascades in Korean spoken question answering (SQA), focusing on downstream semantic failures that conventional ASR metrics cannot fully capture.
Our analysis shows that the relative downstream degradation caused by ASR errors is consistent across LLMs with different absolute performance, suggesting that cascade degradation largely tracks ASR-stage information loss.
We further identify single-character ASR errors as a particularly salient source of information loss in Korean, where even a minimal transcription difference can change the intended question and degrade downstream QA performance.
Finally, an auxiliary comparison shows that a large audio language model outperforms an ASR--LLM cascade with an approximately matched language backbone in noisy Korean SQA, indicating the potential of direct audio input to mitigate transcript-induced information loss.
\end{abstract}

\section{Introduction}

Large language models (LLMs) are increasingly used as general-purpose task solvers for applications such as question answering, information retrieval, and task-oriented dialogue~\cite{zhao2023survey}.
As these applications move into user-facing systems, speech has become an important interaction modality, as seen in speech-based chatbots and voice assistants~\cite{chen2017survey, mctear2002spoken}.
A common implementation is therefore an ASR--LLM cascade, where automatic speech recognition (ASR) first converts the user's speech into text and the resulting transcript is then passed to an LLM as task input~\cite{ji2024wavchat}. 
While this cascade is simple, modular, and easy to deploy, downstream performance becomes dependent on the information preserved in the ASR transcript.

Although modern ASR systems have achieved strong transcription accuracy, their outputs can still degrade under noisy acoustic conditions~\cite{gong1995speech}.
In ASR--LLM cascades, the resulting transcript errors are not merely transcription mistakes, because they can remove or distort the task-relevant information needed by the downstream model~\cite{faruqui2022revisiting}.
This creates a mismatch between transcription-level error rates and downstream task risk, where an error that appears minor in the transcript can still trigger a complete task failure.

Recent work has begun to evaluate ASR outputs by their downstream impact in LLM-powered applications~\cite{pulikodan2025aer, faruqui2022revisiting}.
However, most existing analyses focus on English, leaving language-specific behavior in ASR--LLM cascades underexplored.
Korean is a particularly important case because many Sino-Korean morphemes are realized as single syllables, so character-level ASR errors can be semantic rather than merely orthographic, limiting the direct generalization of English-centered findings.

We address this gap by quantitatively analyzing how ASR errors propagate to downstream QA performance in Korean spoken question answering (SQA).
To obtain a controlled range of ASR error levels, we synthesize speech from Korean text questions, apply additive noise to the synthesized speech, transcribe the resulting audio with ASR, and provide the transcripts to Korean-capable instruction-tuned LLMs together with the original context passage.
We examine downstream QA degradation, test whether an ASR-aware disclaimer prompt mitigates noisy-input errors, and analyze single-character ASR errors on a per-case basis.

The main contributions of this work are summarized as follows:
\begin{itemize}
    \item We show that, in Korean SQA, the relative downstream degradation caused by ASR errors is consistent across LLMs with different absolute performance, suggesting that degradation in ASR--LLM cascades is primarily associated with ASR-stage information loss.
    \item We identify single-character ASR errors as an important source of downstream information loss in Korean ASR--LLM cascades, where even a minimal transcription difference can change the intended question and degrade downstream QA performance.
    \item We show that a direct-audio language model outperforms an ASR--LLM cascade with an approximately matched language backbone in noisy Korean SQA, suggesting that bypassing ASR can reduce transcript-induced information loss.
\end{itemize}

\section{Experimental Setup}
\begin{figure}[t]
    \begin{center}
        \includegraphics[width=70mm]{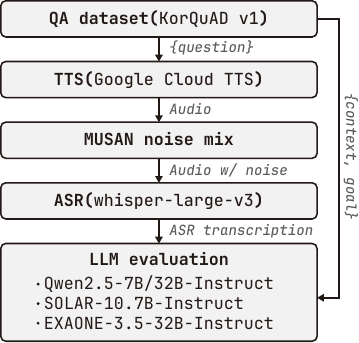}
        \caption{Pipeline for constructing noisy Korean spoken QA inputs and evaluating ASR--LLM performance.}
        \label{fig:pipeline}
    \end{center}
\end{figure}

In this section, we describe the experimental setup for analyzing how ASR errors propagate to downstream Korean SQA. We synthesize Korean questions and mix noise at seven signal-to-noise ratio (SNR) levels to induce a broad range of ASR error rates, yielding an observed character error rate (CER) range of approximately 0.03--0.50. Each noisy question is transcribed by ASR, and the transcript is provided to an LLM as the question, together with the original Korean context passage, for downstream QA evaluation. The full pipeline is illustrated in Fig.~\ref{fig:pipeline}.

\subsection{SQA Evaluation Dataset}

We construct a SQA evaluation dataset from KorQuAD v1~\cite{lim2019korquad}. Starting from the validation split of 5,774 samples, we apply three filtering criteria: questions must be between 5 and 100 characters in length, questions containing special characters are removed, and questions containing digits (0--9) are excluded to avoid ambiguity from digit-to-Hangul conversion during TTS. After filtering, 4,138 candidates remain, from which we select 1,500 samples by stratified sampling to form the final evaluation set. The selected question texts are then synthesized using Google Cloud TTS~\cite{google_tts} with the ko-KR-Wavenet-A Korean voice at 16~kHz mono, where each spoken question is paired with its original text context and gold answer.

We then construct noisy speech inputs by mixing the synthesized questions with noise clips sampled from the noise subset of MUSAN~\cite{snyder2015musan}. The sampled noise clips are mixed with the synthesized speech at seven SNR levels from $+20$ to $-10$~dB in 5~dB steps, in addition to clean conditions.

\subsection{Models and Inference Details}

We use Whisper-large-v3~\cite{radford2023whisper} as the ASR system, using the pretrained model without fine-tuning. 
All speech inputs are decoded in Korean transcription mode using FP16 inference with a batch size of 16. 
For downstream SQA, we compare four instruction-tuned LLMs: Qwen2.5-7B-Instruct, Qwen2.5-32B-Instruct~\cite{qwen25}, SOLAR-10.7B-Instruct~\cite{kim2024solar}, and EXAONE-3.5-32B-Instruct~\cite{exaone35}. 
EXAONE-3.5-32B-Instruct and Qwen2.5-32B-Instruct are run with 4-bit AWQ quantization, whereas SOLAR-10.7B-Instruct and Qwen2.5-7B-Instruct are run in half precision. 
All inferences including ASR and LLM are performed on a single NVIDIA RTX~4090 GPU. 
We run ASR inference with PyTorch~2.4 and perform LLM inference using vLLM~\cite{kwon2023vllm}. 

\begin{figure}[t]
\centering
\includegraphics[width=\linewidth]{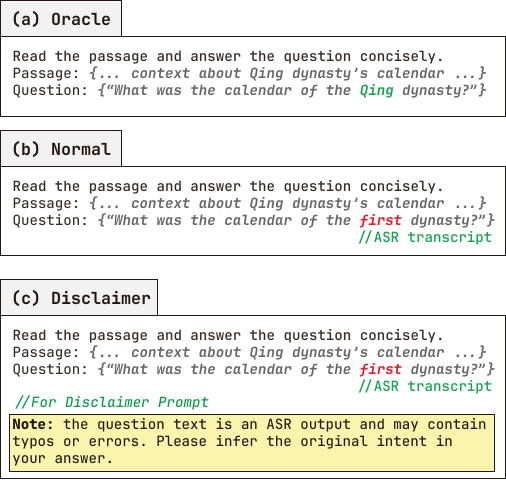}    
    \caption{Example prompts for each condition: \textbf{(a)} \emph{Oracle}, \textbf{(b)} \emph{Normal}, and \textbf{(c)} \emph{Disclaimer}. The example passage, question, and ASR transcript are translated from Korean into English for readability.}
\label{fig:prompt_examples}
\end{figure}

We evaluate each LLM under three prompt conditions. 
In the \textit{Oracle} condition, the original text question is provided to the LLM together with the context passage, bypassing the ASR stage. 
It serves as a reference that approximates the upper bound under perfect transcription.
In the \textit{Normal} condition, the ASR transcript is used as the question input in the standard QA prompt. 
In the \textit{Disclaimer} condition, the same ASR transcript and context passage are used, but the following disclaimer sentence is appended to the QA prompt:
\begin{quote}
``Note: the question text is an ASR output and may contain typos or errors. Please infer the original intent in your answer.''
\end{quote}
Thus, the \textit{Normal} and \textit{Disclaimer} conditions differ only in the presence of this additional prompt sentence.
Example prompts for each condition are shown in Fig.~\ref{fig:prompt_examples}.
Following~\cite{zhang2023moqagpt}, we apply the same LLM-based answer re-extraction step to all generated responses to obtain concise final answer spans, using EXAONE-3.5-32B-Instruct.

\subsection{Metrics}

We use character error rate (CER) instead of word error rate (WER) as the ASR metric, since CER is less affected by Korean word segmentation ambiguity~\cite{park2024kmsav}.
Downstream QA performance is evaluated using exact match (EM) and F1 score, following the standard KorQuAD evaluation protocol.

\section{Results and Analyses}
\subsection{SQA Performance under ASR Corruption}
Table~\ref{tab:mainresult} reports SQA performance in the \textit{Oracle} and \textit{Normal} conditions across clean and noisy speech inputs.
The induced ASR error level remains below 0.07 down to $+5$~dB, but rises sharply under severe noise, from 0.26 at $-5$~dB to 0.50 at $-10$~dB.
Downstream QA performance follows the same degradation pattern. 
Despite differences in \textit{Oracle} performance, all LLMs show similar baseline-relative F1 recovery: about $99\%$ at $+20$~dB, $96\%$ at $+5$~dB, and $67\%$ at $-10$~dB. 
These results suggest that low-error transcripts are nearly sufficient for downstream QA, whereas heavily corrupted transcripts impose a bottleneck that is only weakly mitigated by stronger downstream LLMs.

\begin{table*}[!t]
    \centering
    \caption{QA performance of ASR--LLM cascades under decreasing SNR. Values in parentheses denote the ASR CER for each ASR input condition. Oracle condition uses the original text question and therefore has no CER.}
    \label{tab:mainresult}
    \begin{threeparttable}
    \small
    \setlength{\tabcolsep}{4.2pt}
    \begin{tabular}{llccccccccc}
        \hline
        \multirow{2}{*}{Model} & \multirow{2}{*}{Metric}
        & 
        & \multicolumn{8}{c}{ASR input condition} \\
        \cline{4-11}
        & & \shortstack{\textit{Oracle} \\ \phantom{(0.0315)}}
        & \shortstack{\textit{Clean}\\(0.0315)}
        & \shortstack{+20 dB\\(0.0348)}
        & \shortstack{+15 dB\\(0.0390)}
        & \shortstack{+10 dB\\(0.0473)}
        & \shortstack{+5 dB\\(0.0690)}
        & \shortstack{0 dB\\(0.1143)}
        & \shortstack{$-$5 dB\\(0.2577)}
        & \shortstack{$-$10 dB\\(0.4977)} \\
        \hline
        \multirow{2}{*}{Qwen2.5-7B-Instruct}
        & EM & 0.729 & 0.723 & 0.723 & 0.721 & 0.715 & 0.705 & 0.678 & 0.590 & 0.456 \\
        & F1 & 0.819 & 0.813 & 0.811 & 0.811 & 0.806 & 0.795 & 0.761 & 0.666 & 0.518 \\
        \hline
        \multirow{2}{*}{SOLAR-10.7B-Instruct}
        & EM & 0.569 & 0.558 & 0.559 & 0.555 & 0.559 & 0.545 & 0.533 & 0.471 & 0.393 \\
        & F1 & 0.663 & 0.652 & 0.654 & 0.651 & 0.652 & 0.636 & 0.623 & 0.556 & 0.469 \\
        \hline
        \multirow{2}{*}{Qwen2.5-32B-Instruct}
        & EM & 0.754 & 0.747 & 0.743 & 0.741 & 0.740 & 0.730 & 0.705 & 0.623 & 0.505 \\
        & F1 & 0.849 & 0.841 & 0.839 & 0.837 & 0.835 & 0.824 & 0.799 & 0.710 & 0.580 \\
        \hline
        \multirow{2}{*}{EXAONE-3.5-32B-Instruct}
        & EM & 0.782 & 0.775 & 0.778 & 0.775 & 0.773 & 0.753 & 0.723 & 0.638 & 0.515 \\
        & F1 & 0.870 & 0.863 & 0.865 & 0.863 & 0.860 & 0.839 & 0.808 & 0.718 & 0.586 \\
        \hline
    \end{tabular}
    \end{threeparttable}
\end{table*}

\begin{table}[t]
    \centering
    \caption{Categorization of Korean single-character ASR error cases and their downstream QA performance.}
    \label{tab:onechar}
    \begin{tabular}{lrrc}
        \hline
        Case & \# & Ratio & EM/F1 \\
        \hline
        Single-character ASR errors & 1{,}206 & 100.0\% & 0.534 / 0.647 \\
        \quad Lexical-character errors & 1{,}056 & 87.6\% & 0.532 / 0.643 \\
        \quad Particle-character errors & 150 & 12.4\% & 0.553 / 0.677 \\
        \hline
    \end{tabular}
\end{table}

\subsection{Single-Character ASR Errors as a Salient Source of Information Loss in Korean}

Korean is particularly sensitive to single-character ASR errors because many Sino-Korean morphemes are realized as single syllables, and phonologically similar syllables can correspond to different lexical meanings or syntactic roles. 
Thus, even a one-character substitution can change the intended question.

\begin{figure}[t]
\centering
\includegraphics[width=0.95\linewidth]{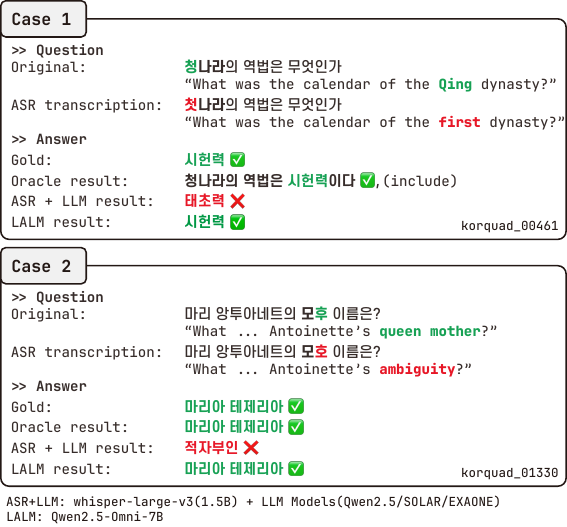}    
    \caption{Representative cases where the ASR--LLM cascade fails due to single-character Korean ASR errors, while the LALM recovers the correct answers.}
\label{fig:onechar-cases}
\end{figure}

Table~\ref{tab:onechar} summarizes 1,206 cases where the reference transcript and the ASR hypothesis differ by exactly one character.
QA performance is computed with Qwen2.5-7B-Instruct for this analysis.
Lexical-character errors are dominant, accounting for 87.6\% of the cases, whereas particle-character errors account for 12.4\%.
Moreover, lexical-character errors result in slightly lower QA performance, suggesting that one-character changes in content words are more likely to disrupt downstream question answering than particle-level changes.
Figure~\ref{fig:onechar-cases} provides representative examples of this phenomenon.

\subsection{LALM vs. ASR--LLM Cascade}

Recent studies have raised the possibility that direct audio input to LLMs can avoid some of the information loss introduced by ASR-based cascades~\cite{hu2024wavllm, choi2025desamo, fathullah2024audiochatllama}, highlighting the potential of large audio language models (LALMs) for downstream speech tasks~\cite{choi2026exploring, peng2026spoken}. 
As an auxiliary comparison, we test whether bypassing ASR with a LALM can reduce the downstream loss observed in ASR--LLM cascades.
We use Qwen2.5-Omni-7B~\cite{xu2025qwenomni} as the representative LALM and evaluate only its Thinker component, whose Qwen2.5-7B-Instruct language backbone allows an approximately controlled comparison with Whisper-large-v3 + Qwen2.5-7B-Instruct.

Table~\ref{tab:lalm} shows that the LALM outperforms the ASR--LLM cascade across all input conditions in this setting, with average gains of $+0.055$ EM / $+0.058$ F1 over the seven noisy SNR conditions.
The gap is already visible at +20\,dB, where ASR corruption is relatively small, and becomes larger under severe noise, increasing to a $+0.112$ F1 gap at $-10$\,dB. 
These results suggest the potential of direct audio input as a way to mitigate ASR-transcript information loss in Korean SQA, while remaining preliminary due to the use of a single LALM.
The cases in Fig.~\ref{fig:onechar-cases} further illustrate this LALM recovery behavior.

\begin{table}[t]
    \centering
    \caption{QA performance of the ASR--LLM cascade and the LALM with an approximately matched language backbone.}
    \label{tab:lalm}
    \begin{tabular}{|r|c|c|c|}
        \hline
        \multicolumn{1}{|c|}{SNR} & \makecell{ASR + LLM \\(EM / F1)} & \makecell{LALM \\(EM / F1)} & \makecell{$\Delta$\\(EM / F1)} \\
        \hline
        \textit{Clean}   & 0.723 / 0.813 & 0.761 / 0.850 & +0.038 / +0.037 \\
        +20 dB  & 0.723 / 0.811 & 0.762 / 0.851 & +0.039 / +0.041 \\
        +15 dB  & 0.721 / 0.811 & 0.764 / 0.851 & +0.043 / +0.040 \\
        +10 dB  & 0.715 / 0.806 & 0.765 / 0.850 & +0.050 / +0.044 \\
        +5 dB   & 0.705 / 0.795 & 0.751 / 0.840 & +0.047 / +0.045 \\
        0 dB    & 0.678 / 0.761 & 0.725 / 0.814 & +0.047 / +0.053 \\
        $-$5 dB  & 0.590 / 0.666 & 0.652 / 0.735 & +0.062 / +0.069 \\
        $-$10 dB & 0.456 / 0.518 & 0.555 / 0.630 & +0.099 / +0.112 \\
        \hline
    \end{tabular}
\end{table} 

\subsection{Effect of Disclaimer Prompting}

Finally, we examine whether the \textit{Disclaimer} condition mitigates ASR-induced QA degradation without changing the ASR system or the downstream LLM.
We compare the \textit{Disclaimer} condition against the \textit{Normal} condition, where the two conditions use the same ASR transcript and context passage and differ only in an additional sentence informing the LLM that the question is an ASR output.
This comparison directly tests whether explicit ASR-error awareness helps recover downstream QA performance.

Table~\ref{tab:disclaimer} compares QA performance under the \textit{Normal} and \textit{Disclaimer} conditions.
Contrary to the expectation that an ASR disclaimer would help the model better handle transcription errors, the disclaimer does not recover noisy QA performance. 
The disclaimer has only a marginal effect on the two 32B-class models, with changes within $\pm 0.006$ EM/F1, while it slightly degrades Qwen2.5-7B and more noticeably degrades SOLAR-10.7B.
These results indicate that disclaimer prompting is not a reliable remedy for ASR degradation, suggesting that improving ASR robustness is a more dependable direction for improving SQA performance.

\section{Conclusion}

We analyzed how ASR errors propagate through ASR--LLM cascades in Korean SQA. 
Our results show that ASR-induced downstream degradation remains similar across LLMs, even though their absolute QA performance differs. 
This suggests that the main bottleneck in this setting is ASR-stage information loss rather than downstream LLM capability.
We also identified single-character ASR errors as a salient source of downstream information loss in Korean SQA: even a minimal transcription difference can change the intended question and degrade downstream QA performance.
Our auxiliary LALM comparison further suggests that direct audio input can mitigate part of the ASR-transcript information loss observed in Korean SQA, although broader validation across LALMs is needed.
This study is limited to Korean SQA based on synthesized speech, and future work should examine whether the findings generalize to broader Korean spoken language understanding tasks and real-speech conditions.

\section*{Acknowledgment}
This research was supported by Culture, Sports and Tourism R\&D Program through the Korea Creative Content Agency grant funded by the Ministry of Culture, Sports and Tourism in 2026 (Project Name: Development of AI-based personalized cultural and arts learning services using smart device, Project Number: RS-2026-25524629).

\begin{table}[t]
    \centering
    \caption{Effect of LLM disclaimer prompting on downstream QA performance, averaged over the seven noisy ASR input conditions from $+20$ to $-10$\,dB.}
    \label{tab:disclaimer}
    \begin{threeparttable}
        \begin{tabular}{|l|c|c|c|c|}
            \hline
            Model & Metric & \textit{Normal} & \textit{Disclaimer} & $\Delta$ \\
            \hline
            \multirow{2}{*}{Qwen2.5-7B}  & EM & 0.6552 & 0.6484 & $-0.0068$ \\
                                         & F1 & 0.7382 & 0.7301 & $-0.0081$ \\
            \hline
            \multirow{2}{*}{SOLAR-10.7B} & EM & 0.5162 & 0.4815 & $-0.0347$ \\
                                         & F1 & 0.6058 & 0.5687 & $-0.0371$ \\
            \hline
            \multirow{2}{*}{Qwen2.5-32B} & EM & 0.6838 & 0.6897 & $+0.0059$ \\
                                         & F1 & 0.7747 & 0.7802 & $+0.0055$ \\
            \hline
            \multirow{2}{*}{EXAONE-3.5-32B}  & EM & 0.7078 & 0.7053 & $-0.0025$ \\
                                         & F1 & 0.7911 & 0.7890 & $-0.0021$ \\
            \hline
        \end{tabular}
    \end{threeparttable}
\end{table}

\printbibliography

\end{document}